\documentclass[a4paper,fleqn]{cas-dc}

\usepackage[authoryear,longnamesfirst]{natbib}

\def\tsc#1{\csdef{#1}{\textsc{\lowercase{#1}}\xspace}}
\tsc{WGM}
\tsc{QE}
\tsc{EP}
\tsc{PMS}
\tsc{BEC}
\tsc{DE}

\begin{document}
\let\WriteBookmarks\relax
\def\floatpagepagefraction{1}
\def\textpagefraction{.001}

\shorttitle{}

\shortauthors{Feng Chen}

\title [mode = title]{CST: Calibration Side-Tuning for Parameter and Memory Efficient Transfer Learning} 




%
\author{Feng Chen}
\cormark[1]


\ead{chenf.edu@gmail.com}


\credit{Conceptualization of this study, Methodology, Software, Data curation, Writing}








\cortext[cor1]{Corresponding author}



\begin{abstract}
Achieving a universally high accuracy in object detection is quite challenging, and the mainstream focus in the industry currently lies on detecting specific classes of objects. However, deploying one or multiple object detection networks requires a certain amount of GPU memory for training and storage capacity for inference. This presents challenges in terms of how to effectively coordinate multiple object detection tasks under resource-constrained conditions.
This paper introduces a lightweight fine-tuning strategy called Calibration side tuning, which integrates aspects of adapter tuning and side tuning to adapt the successful techniques employed in transformers for use with ResNet. The Calibration side tuning architecture that incorporates maximal transition calibration, utilizing a small number of additional parameters to enhance network performance while maintaining a smooth training process. Furthermore, this paper has conducted an analysis on multiple fine-tuning strategies and have implemented their application within ResNet, thereby expanding the research on fine-tuning strategies for object detection networks. Besides, this paper carried out extensive experiments using five benchmark datasets. The experimental results demonstrated that this method outperforms other compared state-of-the-art techniques, and a better balance between the complexity and performance of the finetune schemes is achieved.


\end{abstract}


\begin{highlights}
\item This paper introduces a lightweight fine-tuning strategy called Calibration side tuning, which integrates aspects of adapter tuning and side tuning to adapt the successful techniques employed in transformers for use with ResNet.
\item The Calibration side tuning architecture that incorporates maximal transition calibration, utilizing a small number of additional parameters to enhance network performance while maintaining a smooth training process.
\item This paper has conducted an analysis on multiple fine-tuning strategies and have implemented their application within ResNet, thereby expanding the research on fine-tuning strategies for object detection networks.
\end{highlights}


\maketitle

\section{Introduction} \label{sec1}
Object detection is one of the fundamental problems and key tasks in computer vision, which can be considerable as a target location. It is used to provide the bounding boxes and class labels of visual objects, then the artificial intelligence system can perceive, reason, plan, and act in an object centered manner \cite{tang2021}.

In object detection, deep learning methods are generally improved from R-CNN which was introduced in 2014 \cite{girshick2014}. They first propose candidate boxes, before special classification and optimization localization. In the past period of time, Faster R-CNN \cite{ren2015} can be considered one of the best methods for object detection using a two-stage approach. Yolo \cite{redmon2016} and SSD \cite{liu2016} are equally famous, which are one stage methods that focus more on real-time detection (i.e. target detection time), but have certain shortcomings in accuracy compared to Faster R-CNN.

With the extensive research on Faster R-CNN, scholars have made certain improvements to it by leveraging transformers (such as DETR \cite{carion2020}) and sparse patterns (like Sparse R-CNN \cite{sun2021}), enabling one-stage object detection and enhancing overall detection performance. However, the integration of transformers with CNNs and the introduction of dynamic instance interactive heads have both resulted in increased memory consumption and longer training times, presenting significant challenges for model training.

Object detection has greatly benefited from the advances in supervised deep learning approaches, and the state-of-the-art methods rely on the large-scale fully-annotated datasets \cite{wang2023}. Sufficiently annotated bounding box training data is difficult to obtain due to several reasons: 1) it requires a substantial amount of manual labor, and 2) factors such as economic and dissemination issues contribute to the phenomenon where source data typically follows a long-tailed distribution in the real world \cite{wang2023cut}.

These circumstances heighten the necessity for effective training under low data constraints \cite{fan2021}. In such Few Shot Object Detection (FSOD) problem, the goal is to build detection models for the novel classes with few labeled training images by transferring knowledge from the base classes. FSOD methods can be categorized into meta-learning and fine-tuning approaches \cite{demirel2023}. 

In the face of new object detection tasks, large-scale pre-trained weights and suitable fine-tuning strategies may prove to be effective solutions to address training challenges. In recent years, transfer learning, as a domain adaptation approach \cite{inoue2018}, has been successfully applied in numerous fields such as natural language processing (NLP), computer vision (CV), and vision-and-language (VL) integration.

The transfer learning methods can be divided into four categories: feature-based transfer learning, instance-based transfer learning, relation-based transfer learning, and model-based transfer learning \cite{fan2021dis}. The first three are generally unsupervised transfer learning, which often studies the relationship between the source and target domains and minimizes the differences \cite{li2021}. Meanwhile, more new works try to build generalized detection models that perform well on both base and novel classes. 

The model-based transfer learning is often supervised, and it is called fine tuning. It leverages the weights of large models trained on massive datasets to transfer knowledge from the source domain to the target domain. This robustly generalized knowledge, derived from extensive original datasets, often accelerates model training and enhances its performance.

The general approach taken is that the weights of the first few layers in the original model are frozen, and the back layers are trained. This way is better than training detectors from scratch (full tuning), but the cost is also expensive and may incurs a high risk of over-fitting with the massive parameters of convolutional neural network (CNN) models \cite{wu2020}.

This paper introduces a lightweight fine-tuning strategy called Calibration Side Tuning, which integrates aspects of adapter tuning and side tuning to adapt the successful techniques employed in transformers for use with ResNet. The contributions of this work can be summarized as follows: (1) This paper has conducted an analysis on multiple fine-tuning strategies and have implemented their application within ResNet, thereby expanding the research on fine-tuning strategies for object detection networks. (2) This paper proposes a Calibration side tuning architecture that incorporates maximal transition calibration, utilizing a small number of additional parameters to enhance network performance while maintaining a smooth training process.

The rest of the paper is organized as follows. In Section \ref{sec2}, this paper review related works. In Section \ref{sec3}, this method is described. Section \ref{sec4} illustrates detailed experimental results and comparisons against other state-of-the-art methods. Section \ref{sec5} concludes the paper.





\section{Related works} \label{sec2}
\subsection{Parameter-efficient Transfer Learning}
In practical scenarios, detection frameworks are commonly deployed on mobile or embedded devices, demanding both speed and accuracy. Nevertheless, achieving higher accuracy often entails the use of complex networks that require additional parameters \cite{zhang2022}. Parameter-efficient transfer learning (PETL) is a research direction focused on minimizing the computational cost associated with adapting large pre-trained models to new tasks \cite{sung2022lst}. This is achieved by circumventing updates to the entire set of parameters, leading to an acceleration in training speed. Additionally, PETL facilitates the sharing of parameters across multiple tasks. 

Existing PETL research can be broadly classified into four distinct areas: bit tuning, prompt tuning, adapter tuning, and side tuning. Bit tuning \cite{guo2020, sung2021, zaken2021} involves selecting a sparse subset of parameters from the pre-trained model for updates, without introducing any new parameters. On the other hand, the latter three methods introduce a small number of trainable parameters and exclusively focus on tuning them.

Prompt tuning \cite{lester2021, li2021pre} utilizes a small number of trainable parameters to supplement the input without altering the structure of the pre-trained model. Adapter tuning \cite{houlsby2019, hu2021} introduces stacking or bottleneck modules at a later position within the pre-trained model, thereby modifying the intermediate features. Side-Tuning \cite{zhang2020} employs an additional edge network, often obtained through pruning of the backbone network, and combines the representations from both the edge network and the backbone network to derive the final features.

Current PETL approaches have more focused on image classification \cite{zhao2023} and natural language processing \cite{liu2023}, simply fine tuning in the ResNet or the Transformer can achieve good performance. However, visual object tracking tasks are often carried out by two separate networks working in tandem, making the process of fine-tuning them more challenging \cite{wang2020}. This paper will primarily focus on object detection, aiming to implement a lightweight yet highly effective PETL.

\subsection{Memory-Efficient Training}
Unlike the inference process, model training necessitates back-propagation, hence reverse-mode automatic differentiation requires the preservation of the runtime dataflow graph. During the forward propagation phase, the memory occupied by these intermediate values is not released but rather retained until the back-propagation stage, reaching its peak consumption.

Deep learning frameworks like PyTorch and TensorFlow employ graph optimization and checkpointing techniques to optimize memory usage. Gradient checkpointing \cite{siskind2018} selects to preserve critical computational nodes and discards certain intermediate activations until they are recomputed during back-propagation. The computation graph allows for partial freezing of the model's parameters, enabling it to identify and allocate computational resources only to the genuinely necessary nodes that require updating.

One of the advantages of fine tuning is that it only modifies a small number of parameters, which reduces the time and the required leaf nodes for backpropagation, thereby saving both the time and storage demands for specific adjustments to large models. However, in practical operation, prompt tuning manifests with higher memory consumption compared to full tuning but generally exhibits shorter runtimes than full tuning. Since it is positioned before the backbone network, the intermediate activations in the backbone network must be saved for backpropagation. Bit tuning also suffers from the same issue, whereas adapter tuning and side tuning are less affected by it. The former is typically positioned in the later stages of the backbone network, while the latter operates independently from the backbone network. Consequently, adapter tuning and side tuning prove to be more effective in terms of memory-efficient training.

Furthermore, model compression techniques contribute to memory-efficient training, with notable methods including lightweight network design, parameter quantization, network pruning, and knowledge distillation. The former typically takes place before training, often involving the construction of networks using modules that balance computational cost and effectiveness, such as depth-wise separable convolutions and inverted residuals \cite{howard2017, sandler2018}, or leveraging neural architecture search for network optimization \cite{howard2019}. The latter three methods mainly occur post-training, the goal of quantization is to reduce the accuracy of parameters and intermediate activation maps to low precision \cite{gholami2022}, structure pruning usually drops the less useful filters from a well-trained model for computational efficiency \cite{li2016}, and knowledge distillation generates a lightweight student model based on one or multiple pre-trained cumbersome teacher models \cite{zhang2022con}. This paper will explore the application of lightweight network design and knowledge distillation techniques within fine-tuning to develop more advanced adapter and side networks. 







\section{Proposed Method} \label{sec3}
In Section \ref{sec3.1}, this paper analyzes the application of classical fine-tuning methods on transformers and discuss their applications on ResNet. Subsequently, in Section \ref{sec3.2}, this paper elaborates on a PETL technique called Calibration Side Tuning (CST), along with its architectural specifics. In Section \ref{sec3.3}, network pruning is employed to achieve a more efficient lightweight structure.

\subsection{From Transformer to ResNet} \label{sec3.1}
While not requiring a Region Proposal Network (RPN) as in Faster RCNN, Sparse RCNN features a complex detection head. Both methods, however, incorporate the Feature Pyramid Network (FPN) and Region of Interest (ROI) layers. DETR, on the other hand, adopts a simpler architecture, yet the presence of its encoder, decoder, and detection head still presents unique challenges for fine tuning techniques.

Kaul et al. proposed that RPN is based on its seen classes and updating all parameters in RPN is essential to improve accuracy on novel categories \cite{kaul2022}. FPN which builds multi-scale feature maps is also important for detecting objects. ResNet \cite{he2016} are the backbone of many top-performing systems in several challenging domains \cite{pathiraja2023}. Faster RCNN, Sparse RCNN and DETR utilize it, and thus this paper will explore fine-tuning strategies specifically for ResNet.

Since 2021, vision in transformer (ViT) \cite{dosovitskiy2020} has blurred the lines between natural language processing (NLP) and computer vision (CV), providing a promising path for addressing multi-modal problems. A multitude of fine-tuning techniques have been proposed for transformer-based image detection networks; however, convolutional neural networks (CNNs) still dominate in practical deployments of object detection models. ViTs inherit several advantageous traits from transformers, such as sequence variability and consistent scales across stages, by breaking down images into smaller patches. On the other hand, ResNet presents a more complex structure compared to transformers, which poses significant challenges for fine-tuning methods. This section will primarily focus on adapting some fine-tuning techniques from transformers to make them suitable for application to ResNet.

\subsubsection{Bit Tuning} \label{sec3.1.1}
This approach does not involve adding extra parameters; instead, it investigates which part of the original network is most effective to update. A notable example is BitFit \cite{zaken2021}, which trains only the bias terms and the task-specific classification layer. Since convolutional layers in ResNet do not have bias-terms, this study primarily focuses on the effectiveness of freezing various layers in ResNet for object detection tasks, as well as the performance of networks with train-able bias-terms added atop a frozen ResNet foundation. 

\begin{figure}%
    \centering
    \includegraphics[scale=.75]{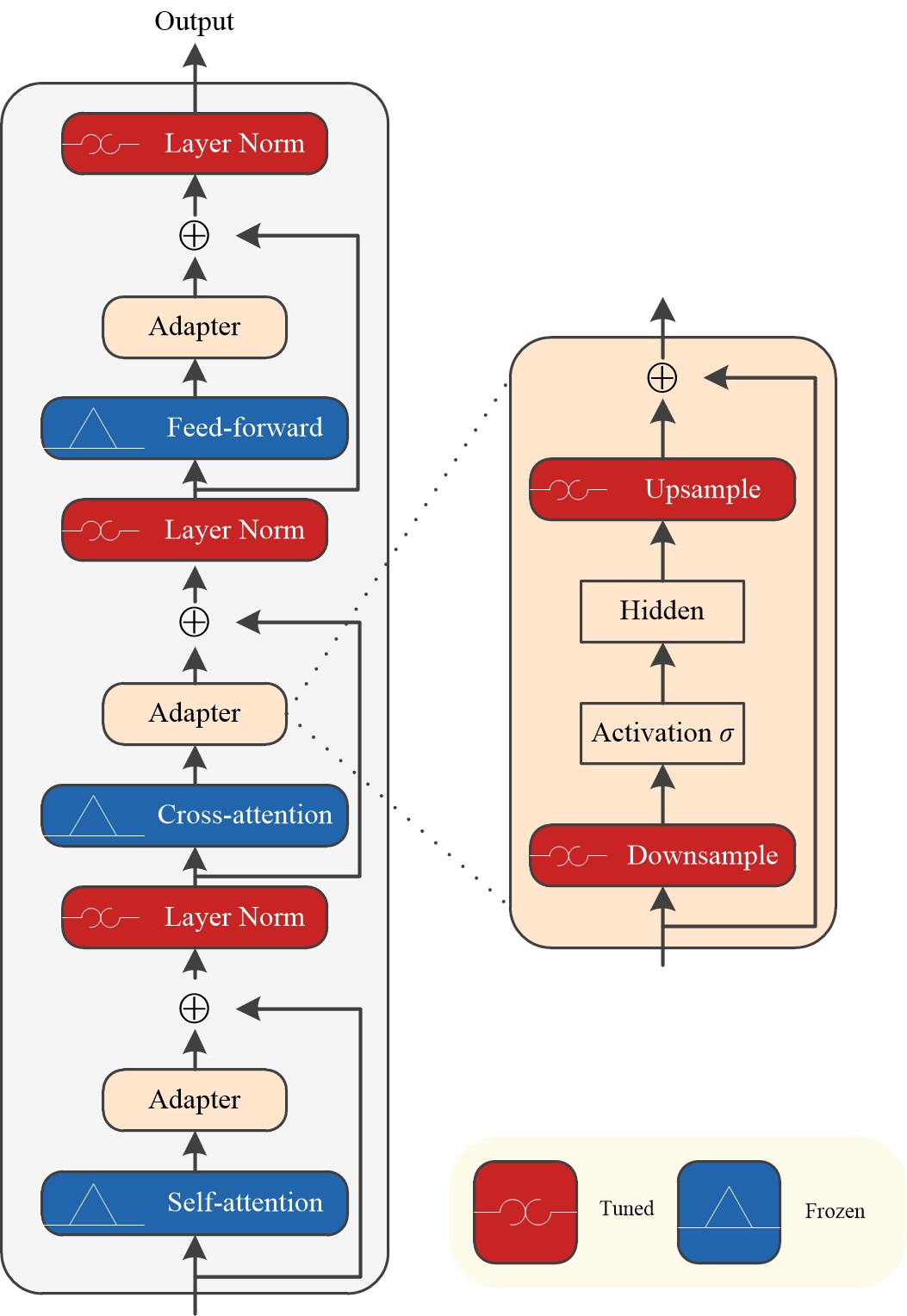}
    \caption{The illustration of VL-Adapter}
    \label{fig1}
\end{figure}

\subsubsection{Adapter Tuning} \label{sec3.1.2}
Adapter Tuning. Adapter tuning can be viewed as a method of composing and reconfiguring existing knowledge by introducing updatable adapters to enhance the performance of the original network. VL-Adapter \cite{sung2022vl} adds adapters after attention blocks and feed-forward layers in the base network, as illustrated in Fig.\ref{fig1}. Considering that ResNet has varying scales and channels across different stages, this paper emulates VL-Adapter by appending the adapter following each stage, which consist of down-sampling and up-sampling convolution pairs with residual connections.

\subsubsection{Prompt Tuning} \label{sec3.1.3}
\begin{figure}%
    \centering
    \includegraphics[scale=.6]{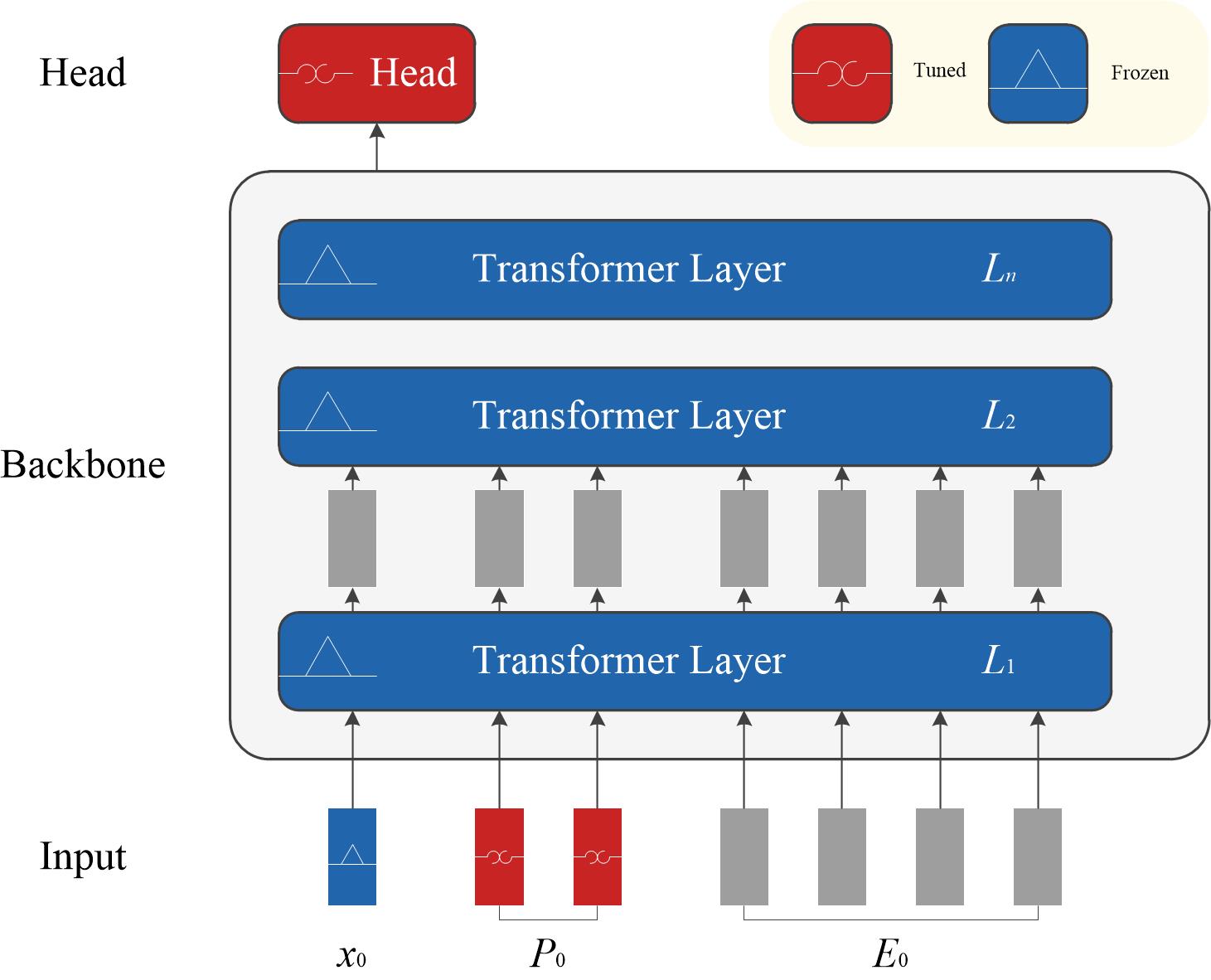}
    \caption{The illustration of VPT}
    \label{fig2}
\end{figure}

Prompt tuning is seen as a method that can help old knowledge understand queries, using a small number of updatable parameters to enrich input information. As shown in Fig.\ref{fig2}, Visual prompt tuning \cite{jia2022} expands the image sequence vector $\emph{E}_0$ with the hint vector $\emph{P}_0$, adapting different tasks with different hint information, making the transformer more generalizable. ResNet does not convert images into multiple embedding vectors like ViT. The features transmitted in the network are three-dimensional and the shape is constantly changing. Due to the difficulty of adding simple information to the input of ResNet, this paper added a convolutional layer before ResNet, similar to performing style conversion on the original image.

\subsubsection{Side Tuning} \label{sec3.1.4}
\begin{figure}%
    \centering
    \includegraphics[scale=.75]{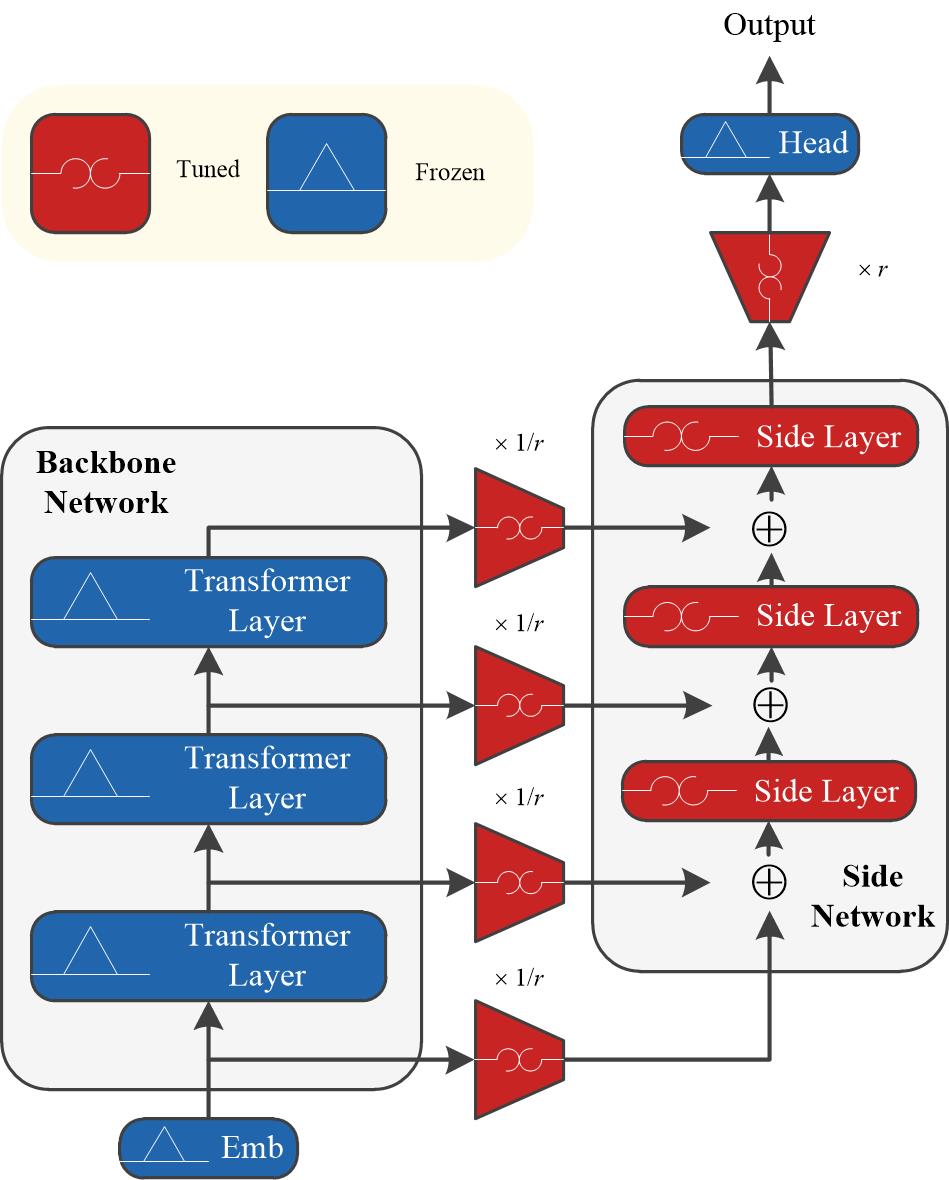}
    \caption{The illustration of Ladder Side-Tuning}
    \label{fig3}
\end{figure}

Side tuning introduces a small auxiliary network that is attached to the original network, with one of its advantages over the previous three methods being that back-propagation only passes through this side network, thereby avoiding the need for temporary storage of activation values from the main network during training. As shown in Fig.\ref{fig3}, Ladder Side-Tuning \cite{sung2022lst} employs channel reduction to minimize memory usage by the side network and subsequently uses a final channel expansion unit to produce output features based on the side network. Given the variations in scale and feature dimensions across ResNet stages, this paper has accordingly adapted the channel reduction units and the structure of the side network.

\subsection{Calibration side tuning} \label{sec3.2}
\begin{figure}%
    \centering
    \includegraphics[scale=.75]{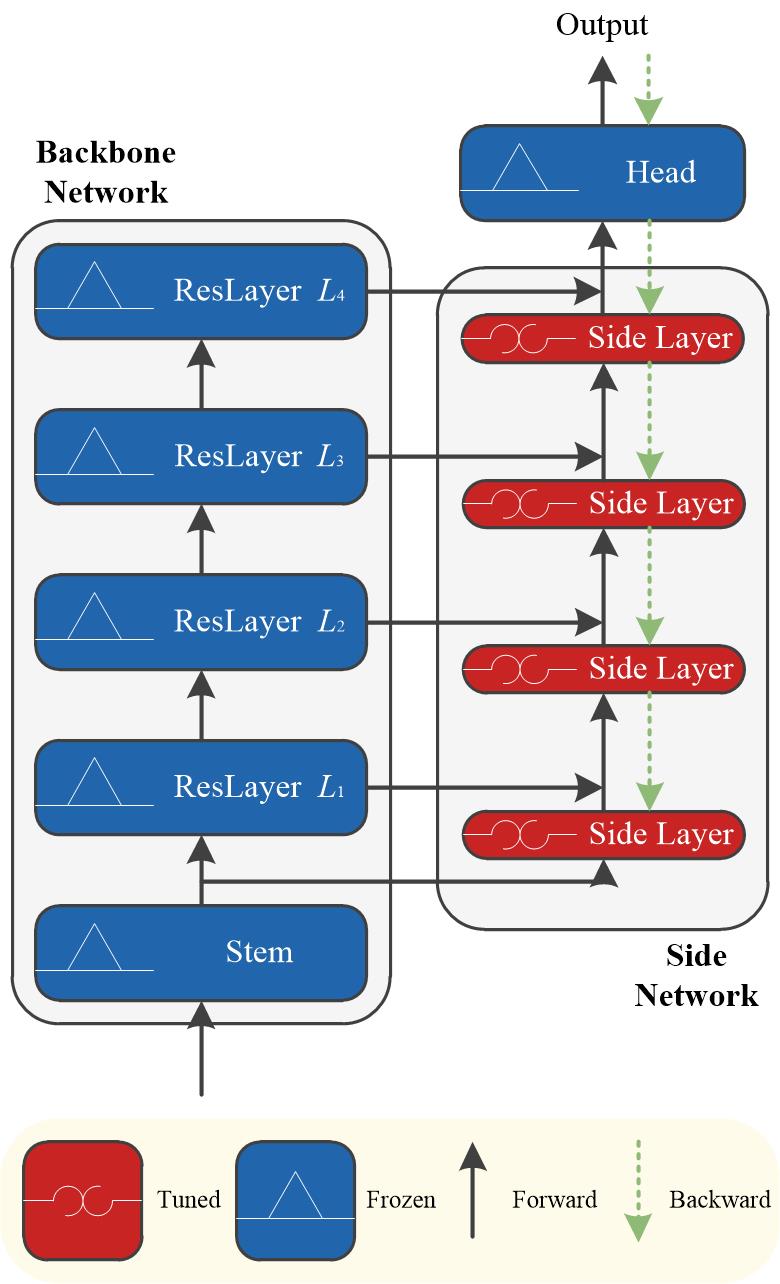}
    \caption{The illustration of Calibration Side-Tuning}
    \label{fig4}
\end{figure}

This paper trains a separate side network, akin to LST, because such independently updatable parameters only require storing the outputs of each stage from the backbone ResNet during back-propagation, as illustrated in Fig.\ref{fig4}. The backbone network contains \emph{n} residual layers ($\emph{L}_1$, $\emph{L}_2$, $\cdot\cdot\cdot$, $\emph{L}_\emph{n}$), and the side network contains \emph{n} side layers ($\emph{S}_1$, $\emph{S}_2$, $\cdot\cdot\cdot$, $\emph{S}_\emph{n}$). In LST, the side network follows the backbone network and can be considered as an additional processing step for features extracted by the backbone network. Unlike LST, this side network operates synchronously with the backbone network. Inspired by Self-Calibrated Net \cite{liu2020}, this paper employs features obtained from the side network to calibrate the backbone network.

The forward propagation of the backbone network is solely determined by itself, as depicted in Eq.\ref{eq1}. The side network initially acquires the output $\emph{L}_0$ from the stem layer and generates $\emph{S}_1$ corresponding to the backbone network's $\emph{L}_1$, as shown in Eq.\ref{eq2}. Subsequently, a gating mechanism is employed to fuse the outputs from both the backbone and side networks, as illustrated in Eq.\ref{eq3}. Furthermore, except for the very first layer of the side network, all subsequent layers take $\emph{G}_\emph{i}$ as their input, as Eq.\ref{eq4}.

\begin{equation}
\emph{L}_{(\emph{i} + 1)} = \emph{F}_{\emph{Res}}^{\emph{i}}\left(\emph{L}_{\emph{i}} \right)\label{eq1}
\end{equation}

\begin{equation}
\emph{S}_{1} = \emph{F}_{\emph{Side}}^{0}\left(\emph{L}_{0} \right)\label{eq2}
\end{equation}

\begin{equation}
\emph{G}_{\emph{i}} = \emph{F}_{\emph{Gate}}^{\emph{i}}\left(\emph{L}_{\emph{i}},\emph{S}_{\emph{i}} \right)\label{eq3}
\end{equation}

\begin{equation}
\emph{S}_{(\emph{i} + 1)} = \emph{F}_{\emph{Side}}^{\emph{i}}\left(\emph{G}_{\emph{i}} \right),i > 0\label{eq4}
\end{equation}

For each 1$\le$ $\emph{i}$ $\le$ $\emph{n}$, $\emph{F}_{\emph{Res}}^{\emph{i}}$$\left(\cdot\right)$denotes the \emph{i}-th residual layer, $\emph{F}_{\emph{Side}}^{0}$$\left(\cdot\right)$ denotes the \emph{i}-th side layer, and $\emph{F}_{\emph{Gate}}^{\emph{i}}$$\left(\cdot,\cdot\right)$ denotes the \emph{i}-th gate module.

\subsubsection{Ladder calibration} \label{sec3.2.1}
The connection between the backbone and side networks can be regarded as a form of gating selection, where mechanisms such as residual connections, concatenation followed by compression mapping, dynamic weights, and branch attentions are common gating strategies. This paper adopts a more lightweight and effective gating mechanism called Maximum Transition Calibration (MTC), as illustrated in Eq.\ref{eq5}.

\begin{equation}
\emph{G}_{\emph{i}} = \emph{F}_{\emph{Gate}}^{\emph{i}}\left(\emph{L}_{\emph{i}},\emph{S}_{\emph{i}} \right) = max\left(\emph{L}_{\emph{i}},\emph{L}_{\emph{i}} \times \emph{S}_{\emph{i}} \right)\label{eq5}
\end{equation}

\subsection{Lightweight design} \label{sec3.3}
The combination of increasingly complex pretrained weights with newly added modules initialized under specific strategies becomes more challenging, and re-searchers generally believe that simply initialized modules may not align well with the overall model. As a result, initialization methods based on knowledge distillation and pruning have emerged. Through empirical validation, it has been found that performing distillation on a specific part of the network might not yield ideal out-comes. In the context of classification tasks, models distilled for these purposes often struggle to perform effectively when transferred to target detection tasks, potentially leading to degraded training results. In the case of using MTC, pruning-based initialization proves to be less effective.

The first two stages of ResNet have coarser feature granularity with stronger generalization capabilities, while the latter two stages possess finer feature granularity with relatively weaker generalization abilities. ResNet's backbone network architecture inherently transforms features from high-resolution and low-channel count to low-resolution and high-channel count, which results in a substantial increase in the number of parameters required for further processing in its third and fourth stages. In this paper, this paper achieves lightweight knowledge fusion through MTC. This paper delves into enhancing the efficiency of CST even more by making the side network more com-pact, specifically by eliminating its intermediate residual layers. 

For the specific construction of the side network, this paper adopts a design similar to VL-Adapter, utilizing pixel-wise convolutions that first shrink and then expand, with internal padding consisting of 3 $\times$ 3 convolutions. Additionally, in consideration of the increasing number of channels layer by layer in ResNet, this paper employs a dynamic scaling factor r to ensure memory usage remains within an appropriate range.

\section{Experiments} \label{sec4}

\subsection{Experimental Settings} \label{sec4.1}
MMDetection \cite{chen2019mm} is an open-source object detection toolbox built upon PyTorch that facilitates the fair implementation of various fine-tuning methods. It provides pre-trained weights trained on MS-COCO \cite{lin2014}. For downstream datasets, this paper utilizes Pascal VOC \cite{everingham2010}, WiderFace \cite{yang2016}, Watercolor \cite{inoue2018}, Comic \cite{inoue2018}, and Crop \cite{KNUST2023}. The first four adhere to the Universal Object Detection Benchmark \cite{bu2021}, while Crop represents a more practical agricultural dataset, which this paper has integrated in the VOC format. This paper reports accuracy using mean Aver-age Precision (mAP), which is calculated using the 11 points interpolation method. Besides, this paper records the maximum memory usage during both training and evaluation for each method.
The L1 loss function and the ADAM optimizer (\emph{$\beta$}1 = 0.9, \emph{$\beta$}2 = 0.999) were used. The training mini-batch size was set to 4, and the size of the output HR image patch was set to 1000$\times$600. The cosine annealing learning scheme is adopted, and the cosine period is 12 epochs. The initial learning rate was set to 2 $\times$ $10^{-3}$. The model was implemented using an NVIDIA 3080 GPU.

\subsection{Ablation analysis} \label{sec4.2}
This paper bases experiments on object detection network Faster RCNN. Initially, this paper conducts an analysis by freezing the layers of ResNet to assess their performance across various datasets. Subsequently, this paper designs different gating mechanisms to substantiate the efficacy of the proposed Maximum Transition Calibration (MTC) technique. This paper compared the performance in mAP, the maximum memory usage during both training and evaluation. 

\subsubsection{Experiments on ResNet} \label{sec4.2.1}

This paper starts from full fine-tuning and incrementally increase the number of frozen layers, with these methods respectively being named as Full, Frozen1, Frozen2, Frozen3, Frozen4. This paper reports the maximum memory usage training on Watercolor for each method. 

In Table \ref{tbl1}, this paper observes in datasets Watercolor, Comic, and Crop that the model performance gradually improves as more residual layers are updated with parameters. This mainly stems from the fact that features trained on base category data lack sufficient discriminative power to classify new instances, leading to a significant drop in performance. Considering both storage consumption and accuracy, freezing stem layers and the first two residual layers prove relatively more efficient. However, in the VOC dataset, the performance of Faster RCNN actually decreases when updating more parameters. On the WiderFace dataset, updating more residual layers does not notably enhance model performance. 

\begin{table*}[]
\caption{The Memory Usage and mAPs of Faster RCNN with different frozen layers}\label{tbl1}
\begin{tabular}{lllllll}
\hline
\multirow{2}{*}{Model} & Memory Usage & \multirow{2}{*}{Watercolor} & \multirow{2}{*}{Comic} & \multirow{2}{*}{WiderFace} & \multirow{2}{*}{VOC} & \multirow{2}{*}{Crop} \\ \cline{2-2}
                       & Train (GB)   &                             &                        &                            &                      &                       \\ \hline
Full                   & 8292         & 0.8965                      & 0.85                   & 0.4459                     & 0.563                & 0.5641                \\
Frozen1                & 6499         & 0.8882                      & 0.8617                 & 0.4452                     & 0.566                & 0.561                 \\
Frozen2                & 5460         & 0.888                       & 0.864                  & 0.4447                     & 0.566                & 0.5609                \\
Frozen3                & 2567         & 0.79                        & 0.6658                 & 0.4434                     & 0.581                & 0.3295                \\
Frozen4                & 2252         & 0.642                       & 0.4655                 & 0.4424                     & 0.589                & 0.2527                \\ \hline
\end{tabular}
\end{table*}

This paper has analyzed this phenomenon and found that: The categories in the VOC dataset are a subset of those in the CoCo dataset, and the pre-trained weights contain knowledge beneficial for detecting its 20 classes. Simple fine-tuning can disrupt these original features, causing performance decline. The WiderFace dataset pertains to face recognition and has a strong correlation with the "person" category in the upstream dataset; fine-tuning adds some information that aids better face detection, but the improvement is not substantial. The Water-color and Comic datasets represent artistic renditions of categories from the up-stream dataset, sharing identical names across six categories but with weaker associations, making it challenging to leverage pre-trained weight knowledge effectively, thus requiring extensive updates. The Crop dataset consists primarily of crop diseases and pests, which are unrelated to categories in the upstream dataset, necessitating greater involvement of residual layers during fine-tuning.

In summary, this paper categorizes downstream datasets into subsets and non-subsets. Non-subsets are further divided into subsets with strong correlations, subsets with weak correlations, and subsets with no correlations. For the subset with weak correlations and the subset without any correlations, updating the latter two residual layers proves to be more effective under resource-constrained conditions.

\subsubsection{Experiments on the gate module} \label{sec4.2.2}
The first experiment concatenates features from the backbone network and the side network, followed by a convolutional layer with a 3×3 kernel size to compress the directly concatenated features, which is referred to as Compress Mapping (CM). The second experiment adopts the dynamic weight scheme from Ladder Side-Tuning, applying a variable weight distribution to the features of both the backbone network and the side network; this method is termed Dynamic Weighting (DW). In the third experiment, the features derived from the side network are passed through a sigmoid function to serve as an attention mechanism that multiplies with the features from the backbone network; this approach is called Branch Attention (BA). Finally, in the fourth experiment, the proposed Maximal Transition Calibration (MTC) technique is utilized to integrate the features from the backbone and side networks.

\begin{table*}[]
\caption{The Memory Usage and mAPs of Faster RCNN with different gating schemes}\label{tbl2}
\begin{tabular}{llllllll}
\hline
\multirow{2}{*}{Model} & \multicolumn{2}{l}{Memory Usage}                                                                                    & \multirow{2}{*}{Watercolor} & \multirow{2}{*}{Comic} & \multirow{2}{*}{WiderFace} & \multirow{2}{*}{VOC} & \multirow{2}{*}{Crop} \\ \cline{2-3}
                       & \begin{tabular}[c]{@{}l@{}}Train   \\ (GB)\end{tabular} & \begin{tabular}[c]{@{}l@{}}Inference \\ (GB)\end{tabular} &                             &                        &                            &                      &                       \\ \hline
CM                     & 3207                                                    & 1695                                                      & 0.5837                      & 0.4363                 & 0.4425                     & 0.572                & 0.2728                \\
DW                     & 2327                                                    & 588                                                       & 0.6318                      & 0.4576                 & 0.4424                     & 0.592                & 0.2468                \\
BA                     & 2345                                                    & 584                                                       & 0.5948                      & 0.4314                 & 0.4426                     & 0.59                 & 0.2402                \\
MTC                    & 2347                                                    & 585                                                       & 0.6758                      & 0.4914                 & 0.4526                     & 0.603                & 0.2599                \\ \hline
\end{tabular}
\end{table*}

In Table \ref{tbl2}, it can be observed that the proposed method MTC outperforms ex-isting basic feature fusion strategies across all datasets. CM, which concatenates and subsequently compresses features using a 3$\times$3 convolutional layer, tends to consume significant storage space and its performance is not consistently superior. However, on the Crop dataset, CM performs better than the other three methods, suggesting that the features learned by the edge network hold some value relative to the back-bone network, and CM might handle inconsistent information more effectively. DW and BA have similar storage requirements, with DW demonstrating better performance compared to BA.

Throughout the training process, this paper finds that the bounding box loss decreases slowly. Faster R-CNN is a two-stage training model, and overly simplistic transformations of the output features may lead to conflicts between new and old knowledge, thus reducing effectiveness. If the new knowledge is simply fused with the old knowledge in a straightforward manner, the new knowledge may “align” too closely with the old, potentially duplicating or reverting to aspects of knowledge that the backbone network has previously discarded. This "knowledge inertia" can cause the new knowledge to drift towards the old, weakening its ability to adapt and up-date. Therefore, the way new and old knowledge are combined requires careful consideration to prevent such an adverse effect.

\subsection{Comparison with state-of-the-art methods} \label{sec4.3}
This paper conducted a comparison with other 4 state-of-the-art methods, which included BitFit (BFT) \cite{zaken2021}, VL-Adapter (VLA) \cite{sung2022vl}, Visual prompt tuning (VPT) \cite{jia2022}, Ladder Side-Tuning (LST) \cite{sung2022lst}. The mAPs on the five test datasets are shown in Table \ref{tbl3}, and memory usage in training and inferencing for each state-of-the-art method are also included.

\begin{table*}[]
\caption{The memory usage and mAPs of Faster RCNN with different fine-tuning methods}\label{tbl3}
\begin{tabular}{llllllll}
\hline
\multirow{2}{*}{Model} & \multicolumn{2}{l}{Memory Usage}                                                                                   & \multirow{2}{*}{Watercolor} & \multirow{2}{*}{Comic} & \multirow{2}{*}{WiderFace} & \multirow{2}{*}{VOC} & \multirow{2}{*}{Crop} \\ \cline{2-3}
                       & \begin{tabular}[c]{@{}l@{}}Train  \\ (GB)\end{tabular} & \begin{tabular}[c]{@{}l@{}}Inference \\ (GB)\end{tabular} &                             &                        &                            &                      &                       \\ \hline
Full                   & 8292                                                   & 579                                                       & 0.8965                      & 0.85                   & 0.4459                     & 0.563                & 0.5641                \\
Frozen4                & 2252                                                   & 579                                                       & 0.642                       & 0.4655                 & 0.4424                     & 0.589                & 0.2527                \\
BFT                    & 7966                                                   & 583                                                       & 0.7154                      & 0.5563                 & 0.4439                     & 0.598                & 0.2554                \\
VPT                    & 7993                                                   & 582                                                       & 0.6575                      & 0.4619                 & 0.4424                     & 0.592                & 0.2481                \\
VLA                    & 6676                                                   & 594                                                       & 0.6698                      & 0.509                  & 0.4428                     & 0.582                & 0.2471                \\
LST                    & 2459                                                   & 582                                                       & 0.355                       & 0.2869                 & 0.4426                     & 0.535                & 0.2408                \\
CST                    & 2347                                                   & 585                                                       & 0.6758                      & 0.4914                 & 0.4526                     & 0.603                & 0.2599                \\ \hline
\end{tabular}
\end{table*}

From Table \ref{tbl3}, it can be seen that applying fine-tuning methods to ResNet leads to a certain degree of performance improvement. On datasets such as Watercolor, Comic, and Crop, which are less related to the upstream datasets, the accuracy is slightly higher than when all parameters of ResNet are frozen, but significantly inferior to full finetuning. In datasets like Pascal VOC and WiderFace, which have stronger associations with subsets, the fine-tuning methods outperform full finetuning. Among the existing fine-tuning techniques, LST performs the worst, while VPT and VLA exhibit similar performance, sometimes surpassing frozen ResNet and at other times underperforming it. BFT and this method achieve higher accuracy than frozen ResNet across all five datasets. On the Watercolor and Comic datasets, BFT outperforms this method; however, during training, this method requires significantly lower storage space compared to BFT. Furthermore, on the WiderFace dataset, BFT's accuracy is lower than full fine-tune, whereas this method's accuracy exceeds that of full finetune.

The difference between old and new knowledge often manifests in strong classification capabilities but weaker bounding box regression abilities, likely due to confusion between the two types of knowledge leading to loss of high-frequency information within features. The low-occupancy edge network in LST struggles to align with the backbone network, causing a "knowledge mismatch" that results in slower reduction of detection box losses and, in severe cases, difficulties in inference. Since feature pyramid networks (FPNs) are commonly employed in object detection tasks to adjust old knowledge, adapter tuning may not yield significant benefits. Prompt tuning is challenging to design for ResNet and thus its effectiveness is limited. BFT simply increases biases in convolutional layers, which can be viewed as a normalization enhancement to the backbone network, thereby improving its performance. This method, on the other hand, utilizes newly learned knowledge to verify features, and the use of maximal transition calibration makes the training process smoother, ultimately enhancing model accuracy.

\section{Conclusions} \label{sec5}
This paper introduces a lightweight fine-tuning strategy called Calibration Side Tuning. This scheme draws upon the successful experiences of adapter tuning and side tuning as applied in transformers, adapting and implementing them on ResNet. Calibration Side Tuning employs maximal transition calibration to effectively integrate features from the backbone network and the edge network. This paper carried out extensive experiments using five datasets. The experimental results demonstrated that this method outperforms other compared state-of-the-art techniques, and the approach provides a better balance between the complexity and performance of the finetune methods. Furthermore, conventional methods generally focus on subset datasets or subsets with strong correlations, while relatively fewer studies have been conducted on weakly correlated and uncorrelated subset datasets. This paper analyzes the performance of fine-tuning strategies across various types of databases and compare different feature fusion approaches. This work provides empirical insights into the application of transfer learning for training and deployment in deep learning networks.



\printcredits

\bibliographystyle{cas-model2-names}
\bibliography{cas-refs}





\end{document}